# TRAPEZOIDAL FUZZY NUMBERS FOR THE TRANSPORTATION PROBLEM

**ARINDAM CHAUDHURI***
*Lecturer (Mathematics & Computer Science)*
*Meghnad Saha Institute of Technology,*
*Kolkata, India*
*arindam_chau@yahoo.co.in*
*\*corresponding author*

**KAJAL DE**
*Professor in Mathematics*
*School of Science*
*Netaji Subhas Open University,*
*Kolkata, India*
*kajalde@rediffmail.com*

**DIPAK CHATTERJEE**
*Distinguished Professor*
*Department of Mathematics*
*St. Xavier's College,*
*Kolkata, India*

**PABITRA MITRA**
*Assistant Professor*
*Department of Computer Science Engineering*
*Indian Institute of Technology,*
*Kharagpur, India*

**Abstract**

Transportation Problem is an important problem which has been widely studied in Operations Research domain. It has been often used to simulate different real life problems. In particular, application of this Problem in NP-Hard Problems has a remarkable significance. In this Paper, we present the closed, bounded and non–empty feasible region of the transportation problem using fuzzy trapezoidal numbers which ensures the existence of an optimal solution to the balanced transportation problem. The multi-valued nature of Fuzzy Sets allows handling of uncertainty and vagueness involved in the cost values of each cells in the transportation table. For finding the initial solution of the transportation problem we use the Fuzzy Vogel's Approximation Method and for determining the optimality of the obtained solution Fuzzy Modified Distribution Method is used. The fuzzification of the cost of the transportation problem is discussed with the help of a numerical example. Finally, we discuss the computational complexity involved in the problem. To the best of our knowledge, this is the first work on obtaining the solution of the transportation problem using fuzzy trapezoidal numbers.

*Keywords*: Transportation Problem, Linear Programming Problem, Fuzzy trapezoidal numbers, Fuzzy Vogel's Approximation Method, Fuzzy Modified Distribution Method

## 1. Introduction

The transportation problem is a special type of the linear programming problem. It is one of the earliest and most fruitful applications of linear programming technique. It has been widely studied in Logistics and Operations Management where distribution of goods and commodities from sources to destinations is an important issue[10]. The origin of the transportation methods dates back to 1941 when F. L. Hitchcock[7] presented a study entitled *The Distribution of a Product from Several Sources to Numerous Localities*. This presentation is considered to be the first important contribution to the solution of the Transportation Problems. In 1947 T. C. Koopmans[11] presented an independent study called *Optimum Utilization of the Transportation System*. These two contributions helped in the development of transportation methods which involve a number of shipping sources and a number of destinations. An earlier approach was given by Kantorovich[10]. The linear programming formulation and the associated systematic method for solution were first given by G. B. Dantzig[4]. The computational procedure is an adaptation of the simple method applied to the system of equations of the associated linear programming problem. The task of distributor's decisions can be optimized by reformulating the distribution problem as generalization of the classical transportation problem. The conventional transportation problem can be represented as a mathematical structure which comprises an objective function subject to certain constraints. In classical approach, transporting costs from *m* sources or wholesalers to the *n* destinations or consumers are to be minimized. It is an optimization problem which has been applied to solve various NP-Hard problems.

Within a given time period each shipping source has a certain capacity and each destination has certain requirements with a given cost of shipping from the source to the destination. The objective function is to minimize total transportation costs and satisfy the destination requirements within the source requirements[8, 13]. However, in real life situations, the information available is of imprecise nature and there is an inherent degree of vagueness or uncertainty present in the problem under consideration. In order to tackle this uncertainty the concept of Fuzzy Sets can be used as an important decision making tool[15]. Imprecision here is meant in the sense of vagueness rather than the lack of knowledge about the parameters present in the system. The Fuzzy Set Theory thus provides a strict mathematical framework in which vague conceptual phenomena can be precisely and rigorously studied[3].

In this work, we discuss the balanced transportation problem and related theorems which describe important mathematical characteristics, and then develop in terms of the revised simplex method computational procedure for solving the problem. Though this problem can be solved by using the simplex method, its special structure allows us to develop a simplified algorithm for its solution. This model is not representative of a particular situation but may arise in many physical situations. Considering this point of view we develop the model using trapezoidal fuzzy numbers[1, 2]. The nature of the solution is closed, bounded and non–empty feasible which ensures the existence of an optimal solution to the balanced transportation problem. The cost values of each cell in the transportation table are represented in terms of the trapezoidal fuzzy numbers which allows handling of uncertainty and vagueness involved. The initial solution of the transportation problem is calculated using the Fuzzy Vogel's Approximation Method and optimality test of the solution is performed using Fuzzy Modified Distribution Method.

This paper is organized as follows. In section 2, the trapezoidal membership function is defined. In the next section, the general transportation problem with fuzzy trapezoidal numbers is discussed. This is followed by the solution of transportation problem using fuzzy trapezoidal numbers in section 4. Section 5 illustrates the solution of transportation problem through a numerical example. The computational complexity of the problem is given in section 6, followed by discussions in section 7. Finally, in section 8 conclusions are given.

## 2. Trapezoidal Membership Function

The trapezoidal membership function[14] is specified by four parameters $\{a, b, c, d\}$ as follows:

$$trapezoid(x;a,b,c,d) = \begin{cases} 0, x < a \\ (x-a)/(b-a), a \leq x < b \\ 1, b \leq x < c \\ (d-x)/(d-c), c \leq x < d \\ 0, x \geq d \end{cases} \quad (1)$$

The figure 1 below illustrates an example of a trapezoidal membership function defined by *trapezoid*(*x*; 10, 20, 60, 96).

## 3. Transportation Problem with Fuzzy Trapezoidal Numbers

Assume a situation having $m$ origins or supply centers which contain various amounts of commodity that has to be allocated to $n$ destinations or demand centers. Consider that the $i^{th}$ origin must supply the fuzzy quantity $A_i = [a_i^{(1)}, a_i^{(2)}, a_i^{(3)}, a_i^{(4)}] (> [-\delta, 0, 0, \delta])$, whereas the $j^{th}$ destination must receive the fuzzy quantity $B_j = [b_j^{(1)}, b_j^{(2)}, b_j^{(3)}, b_j^{(4)}] (> [-\delta, 0, 0, \delta])$. Let the fuzzy cost $C_{ij} = [c_{ij}^{(1)}, c_{ij}^{(2)}, c_{ij}^{(3)}, c_{ij}^{(4)}]$ of shipping a unit quantity from the origin $i$ to the destination $j$ be known for all the origins $i$ and destinations $j$. As it is possible to transport from any one origin to any one destination, and the problem is to determine the number of units to be transported from origin $i$ to the destination $j$ such that all requirements are satisfied at a total minimum transportation cost. This scenario holds for a *balanced transportation problem*. Further in an *unbalanced transportation problem*[5, 12] the sum availabilities or supplies of the origins are not equal to the sum of the requirements or demands at the destinations. In order to solve this problem we first convert the unbalanced problem into a balanced one by artificially converting it to a problem of equal demand and supply. For that, we introduce a fictitious or dummy origin or destination that will provide the required supply or demand respectively. The costs of transporting a unit from the fictitious origin as well as the costs of transporting a unit to the fictitious destination are taken as zero. This is equivalent to not transporting from a dummy source or to a dummy destination with zero transportation cost.

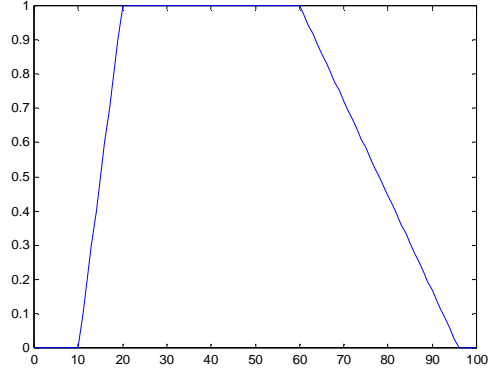

**Figure 1: Trapezoidal membership function defined by *trapezoid*(*x*; 10, 20, 60, 96)**

The mathematical formulation of the problem is as follows. Let $X_{ij} = [x_{ij}^{(1)}, x_{ij}^{(2)}, x_{ij}^{(3)}, x_{ij}^{(4)}]$ be the fuzzy number of units supplied from the origin $i$ to the destination $j$. Then the problem can be written as:

$$\text{Minimize } Z = \sum_{i=1}^{m} \sum_{j=1}^{n} C_{ij} X_{ij} \quad (\text{here, } Z = [z^{(1)}, z^{(2)}, z^{(3)}, z^{(4)}]) \tag{2}$$

*subject to constraints*:

$$\sum_{j=1}^{n} X_{ij} = A_i, i = 1, 2, \ldots, m \quad \text{(supply constraints)} \tag{3}$$

$$\sum_{i=1}^{m} X_{ij} = B_j, j = 1, 2, \ldots, n \quad \text{(destination constraints)} \tag{4}$$

$$\sum_{j=1}^{n} X_{ij} \geq [-\delta, 0, 0, \delta] \text{ for all } i \text{ and } j \tag{5}$$

where, $A_i > [-\delta, 0, 0, \delta]$ and $B_j > [-\delta, 0, 0, \delta]$ for all $i$ and $j$. It should be noted that the Transportation Problem is a linear program. Suppose that there exists a feasible solution to the problem. Then, it follows, from equations (3) and (4), that

$$\sum_{i=1}^{m} \sum_{j=1}^{n} X_{ij} = \sum_{i=1}^{m} A_i = \sum_{j=1}^{n} B_j$$

Thus, for the problem to be consistent, we must have the following consistency equation:

$$\sum_{i=1}^{m} A_i = \sum_{j=1}^{n} B_j \tag{6}$$

If the problem is inconsistent, the following equation holds:

$$\sum_{i=1}^{m} A_i \neq \sum_{j=1}^{n} B_j \tag{7}$$

If the consistency condition (6) holds, then the transportation problem is *balanced*, such that the total supply is equal to the total demand otherwise the problem is *unbalanced*. It is obvious from the constraints (3), (4) and (5) that every component $X_{ij}$ of a fuzzy feasible solution vector $X$ is bounded, i.e.,

$$[-\delta, 0, 0, \delta] \leq [x_{ij}^{(1)}, x_{ij}^{(2)}, x_{ij}^{(3)}, x_{ij}^{(4)}] \leq \min([a_i^{(1)}, a_i^{(2)}, a_i^{(3)}, a_i^{(4)}], [b_j^{(1)}, b_j^{(2)}, b_j^{(3)}, b_j^{(4)}])$$

Thus, the feasible region of the problem is closed, bounded and non-empty. Hence, there always exists an optimal solution to the balanced transportation problem. Constraint equations (3) and (4) can be written in the matrix form as follows:

$$AX = B$$

with $X = (X_{11}, X_{12}, ...., X_{1n}, X_{21}, ...., X_{2n}, .... X_{mn})^T$,

$B = (A_1, A_2, ...., A_m, B_1, B_2, ...., B_n)^T$ and $A$ as an $(m + n) \times mn$ matrix given by

$$A = \begin{bmatrix} 1 .................... 0 \\ 0..1 ............... 0 \\ .................... . \\ .................... . \\ 0 .................... 1 \\ I..I..I..I..I..I...I \end{bmatrix}$$

Here, 1 is the $1 \times n$ matrix with all the components as 1 and $I$ is the n × n identity matrix. Since the sum of *m* equations (3) equals the sum of *n* equations (4), the (*m* + *n*) rows of *A* are linearly dependent. This implies that *rank* $(A) \leq m + n - 1$.

The transportation model has a special structure which enables us to represent it in the form of rectangular array called the *transportation table* as given in figure 2. In this table, each of the *mn* cells corresponds to a variable; each row corresponds to one of the *m* constraints (3) called row constraints and each column corresponds to one of the *n* constraints (4) called column constraints. The $(i, j)^{th}$ cell at the intersection of the $i^{th}$ row and $j^{th}$ column contains cost $C_{ij}$ and decision variable $X_{ij}$. The cells in the transportation table can be classified as occupied cells and unoccupied cells. The allocated cells in the transportation table are called occupied cells and empty cells are called unoccupied cells.

### 3.1 Definitions: *Fuzzy feasible solution of the Transportation Problem*

Now we give some important definitions relevant for developing the feasible solution for the transportation problem using the trapezoidal fuzzy numbers. They are briefly enumerated as follows:

a) **Fuzzy feasible solution:** Any set of fuzzy non-negative allocations ($X_{ij} > [-\delta, 0, 0, \delta]$, $\delta$ being a small positive number) which satisfies the row and column sum is a fuzzy feasible solution.

b) **Fuzzy basic feasible solution:** A fuzzy feasible solution is a fuzzy basic feasible solution if the number of non-negative allocations is at most $(m + n -1)$; where, $m$ is the number of rows, $n$ is the number of columns in the transportation table.

c) **Fuzzy non–degenerate basic feasible solution:** Any fuzzy feasible solution to the transportation problem containing $m$ origins and $n$ destinations is said to be fuzzy non– degenerate, if it contains exactly $(m + n - 1)$ occupied cells.

d) **Fuzzy degenerate basic feasible solution:** If the fuzzy basic feasible solution contains less than $(m + n - 1)$ non-negative allocations, it is said to be fuzzy degenerate.

**DESTINATIONS**

|  | $D_1$ | $D_2$ | ------- | $D_n$ |  |
|---|---|---|---|---|---|
| $O_1$ | $C_{11}$ | $C_{12}$ | ------- | $C_{1n}$ | $A_1$ |
| $O_2$ | $C_{21}$ | $C_{22}$ | ------- | $C_{2n}$ | $A_2$ |
| ⋮ | ------- | ------- | ------- | ------- | ⋮ |
| $O_m$ | $C_{m1}$ | $C_{m2}$ | ------- | $C_{mn}$ | $A_m$ |
|  | $B_1$ | $B_2$ | ------- | $B_n$ |  |

(ORIGINS on left side)

**Figure 2: Transportation Table**

### 3.2 *Theorems*

Finally, we present some important theorems often found useful while developing the feasible solution for the transportation problem using the trapezoidal fuzzy numbers as well as testing the optimality of the obtained solution. The proofs of these theorems can be found in[5, 6, 12].

**Theorem 1:** *The number of basic variables in a transportation problem is at most $(m + n – 1)$.*

**Theorem 2:** *The transportation problem always has a feasible solution.*

**Theorem 3:** *All bases for the transportation problem are triangular (upper or lower) in nature.*

**Theorem 4:** *The values of the basic variables in a basic feasible solution to the transportation problem are given by the expressions of the form*

$$X_{ij} = \pm \sum_{some\,p} A_p \mp \sum_{some\,q} B_q \quad (8)$$

*where, (in $\pm$ and $\mp$) the upper signs apply to some basic variables and the lower signs apply to the remaining basic variables.*

**Theorem 5:** *The solution of the transportation problem is never unbounded.*

**Theorem 6:** *A subset of the columns of the coefficient matrix of a transportation problem is linearly dependent, if and only if, the corresponding cells or a subset of them can be sequenced to form a loop.*

**Theorem 7:** *If there be a feasible solution having $(m + n - 1)$ independent positive allocations and if there be numbers $u_i$ and $v_j$, $(i = 1,............,m; j = 1,..............,n)$ satisfying $c_{rs} = u_r + v_s$ for each occupied cell $(r, s)$, then the cell evaluation $\Delta_{ij}$ corresponding to the unoccupied cell $(i, j)$ will be given by $\Delta_{ij} = c_{ij} - (u_i + v_j)$.*

### 4. Solution of the Transportation Problem using Fuzzy Trapezoidal Numbers

The solution of the fuzzy transportation problem is generally obtained in following two stages:

a) Initial basic feasible solution

b) Test of optimality for the solution

#### 4.1 *Fuzzy Vogel's Approximation Method (FVAM)*

The initial basic feasible solution can be easily obtained using the methods like North West Corner Rule, Least Cost Method or Matrix Minima Method, Vogel's Approximation Method (VAM) etc. VAM is preferred over the other methods, since the initial basic feasible solution obtained by this method is either optimal or very close to the optimal solution. Here, we discuss only VAM and using the fuzzy trapezoidal numbers. The steps involved in FVAM for finding the fuzzy initial solution are briefly enumerated below[12]:

**Step 1:** The penalty cost is found by considering the difference the smallest and next smallest costs in each row and column.

**Step 2:** Among the penalties calculated in **step 1**, the maximum penalty is chosen. If the maximum penalty occurs more than once then any one can be chosen arbitrarily.

**Step 3:** In the selected row or column found in **step 2**, the cell having the least cost is considered. An allocation is made to this cell by taking the minimum of the supply and demand values.

**Step 4:** Finally, the row or column is deleted which is fully fuzzy exhausted. Now, considering the reduced transportation tables repeat **steps 1 - 3** until all the requirements are fulfilled.

#### 4.2 *Fuzzy Modified Distribution Method (FMODIM)*

Once the fuzzy initial basic feasible solution has been obtained, the next step is to determine whether the solution obtained is fuzzy optimum or not. Optimality test can be conducted to any initial basic feasible solution of the transportation problem provided such allocations have exactly $(m + n - 1)$ non–negative allocations, where *m* is the number of origins and *n* is the number of destinations. Also these allocations must be in independent positions. To perform the optimality test, we make use of the FMODIM using the fuzzy trapezoidal numbers. The various steps involved in FMODIM for performing the optimality test are given below[12]:

**Step 1:** Find the fuzzy initial basic feasible solution of the fuzzy transportation problem by using FVAM.

**Step 2:** Find a set of numbers $U_i = [u_i^{(1)}, u_i^{(2)}, u_i^{(3)}, u_i^{(4)}]$ and $V_j = [v_j^{(1)}, v_j^{(2)}, v_j^{(3)}, v_j^{(4)}]$ for each row and column satisfying $U_i (+) V_j = C_{ij}$ for each occupied cell. We start by assigning a number *fuzzy zero* (which may be [-0.05, 0, 0, 0.05]) to any row or column having the maximum number of allocations. If the maximum number of allocation is more than one, we choose any one arbitrarily.

**Step 3:** For each empty or unoccupied cell, we find the sum of $U_i$ and $V_j$ and write it in each cell.

**Step 4:** Find the net evaluation value for each empty cell given as, $\Delta_{ij} = C_{ij}(-)(U_i(+)V_j)$ and also write it in each cell. This gives the optimality conclusion which may be any of the following:

a) If all $\Delta_{ij} > [-\delta, 0, 0, \delta]$, the solution is fuzzy optimum and a fuzzy unique solution exists.

b) If $\Delta_{ij} \geq [-\delta, 0, 0, \delta]$, then the solution is fuzzy optimum, but an alternate solution exists.

c) If at least one $\Delta_{ij} < [-\delta, 0, 0, \delta]$, the solution is not fuzzy optimum. In this case we go to the next step, to improve the total transportation cost.

**Step 5:** Select the empty cell having the most negative value of $\Delta_{ij}$. From this cell we draw a closed path by drawing horizontal and vertical lines with the corner cells occupied. Assign positive and negative signs alternately and find the minimum allocation from the cell having the negative sign. This allocation is to be added to the allocation having positive sign and subtracted from the allocation having negative sign.

**Step 6:** The **step 5** yields a better solution by making one or more occupied cell as empty and one empty cell as occupied. For this new set of basic feasible allocations repeat from **steps 2 – 5** until an optimum basic feasible solution is obtained.

### 5. Numerical Example

In this section, we consider the fuzzy transportation problem and obtain the fuzzy initial basic feasible solution of the problem by FVAM and determine the fuzzy membership functions of costs and allocations. We then test the optimality of the solution obtained using FMODIM.

#### 5.1 *Initial Basic Feasible Solution by FVAM*

The transportation problem consists of 3 origins and 4 destinations. The cost coefficients are denoted by trapezoidal fuzzy numbers. The corresponding availability (supply) and requirement (demand) vectors are also given in the figure below. In the above Table, $\sum_i A_i = [1009.85, 1010, 1010, 1010.15]$ (column sum) and $\sum_j B_j = [1009.80, 1010, 1010, 1010.20]$ (row sum). Since, $\sum_i A_i$ and $\sum_j B_j$ are fuzzy equal, differing by the fuzzy zero viz., [-0.35, 0, 0, 0.35] the given problem is balanced and there exists a fuzzy feasible solution to the problem. We first find the row and column penalty as the difference between the fuzzy least and next fuzzy least cost in the corresponding rows and columns respectively. In the above problem the maximum penalty is [6.9, 7, 7.1] corresponding to $D_2$ column. In this allocation the cell having the fuzzy least cost is (1, 2). To this cell we allocate the minimum of Supply $A_1$ and Demand $B_2$ i.e., ([279.95, 280, 280, 280.05], [249.95, 250, 250, 250.05]) = [249.95, 250, 250, 250.05] as given in Figure 2. This exhausts the second column

by fuzzy zero, [-0.10, 0, 0, 0.10] and supply is reduced to ([279.95, 280, 280, 280.05] (-) [249.95, 250, 250, 250.05]) = [29.90, 30, 30, 30.10]. The second column is deleted from the Figure 2 such that we have the following shrunken matrix given in Figure 3. As the second column is deleted the values of penalties are also changed.

|  |  | DESTINATIONS | | | | |
|---|---|---|---|---|---|---|
|  |  | $D_1$ | $D_2$ | $D_3$ | $D_4$ | Supply ($A_i$) |
| ORIGINS | $O_1$ | [12.95, 13, 13, 13.05] | [14.95, 15, 15, 15.05] | [15.95, 16, 16, 16.05] | [17.95, 18, 18, 18.05] | [279.95, 280, 280, 280.05] |
|  | $O_2$ | [19.95, 20, 20, 20.05] | [21.95, 22, 22, 22.05] | [10.95, 11, 11, 11.05] | [7.95, 8, 8, 8.05] | [329.95, 330, 330, 330.05] |
|  | $O_3$ | [18.95, 19, 19, 19.05] | [24.95, 25, 25, 25.05] | [16.95, 17, 17, 17.05] | [10.95, 11, 11, 11.05] | [399.95, 400, 400, 400.05] |
|  | Demand ($B_j$) | [299.95, 300, 300, 300.05] | [249.95, 250, 250, 250.05] | [279.95, 280, 280, 280.05] | [179.95, 180, 180, 180.05] |  |

**Figure 1: Transportation Problem with 3 Origins and 4 Destinations**

|  |  | DESTINATIONS | | | | | |
|---|---|---|---|---|---|---|---|
|  |  | $D_1$ | $D_2$ | $D_3$ | $D_4$ | Supply ($A_i$) | Row Penalty |
| ORIGINS | $O_1$ | [12.95, 13, 13, 13.05] | **[14.95, 15, 15, 15.05]** ([249.95, 250, 250, 250.05]) | [15.95, 16, 16, 16.05] | [17.95, 18, 18, 18.05] | [279.95, 280, 280, 280.05] | [1.9, 2, 2, 2.1] |
|  | $O_2$ | [19.95, 20, 20, 20.05] | [21.95, 22, 22, 22.05] | [10.95, 11, 11, 11.05] | [7.95, 8, 8, 8.05] | [329.95, 330, 330, 330.05] | [2.9, 3, 3, 3.1] |
|  | $O_3$ | [18.95, 19, 19, 19.05] | [24.95, 25, 25, 25.05] | [16.95, 17, 17, 17.05] | [10.95, 11, 11, 11.05] | [399.95, 400, 400, 400.05] | [5.9, 6, 6, 6.1] |
|  | Demand ($B_j$) | [299.95, 300, 300, 300.05] | [249.95, 250, 250, 250.05] [-0.10, 0, 0, 0.10] | [279.95, 280, 280, 280.05] | [179.95, 180, 180, 180.05] |  |  |
|  | Column Penalty | [5.9, 6, 6, 6.1] | [6.9, 7, 7, 7.1] | [5.9, 6, 6, 6.1] | [2.9, 3, 3, 3.1] |  |  |

**Figure 2: 1st Allocation to the Transportation Problem**

In Figure 3, the maximum value of Penalty is [5.9, 6, 6, 6.1] corresponding to $D_1$ column and $O_3$ row. Taking either of the row or column viz., if we consider row $O_3$; the cell having the least fuzzy cost is (3, 4). To this cell we allocate the minimum of Supply $A_3$ and Demand $B_4$ i.e., ([399.95, 400, 400, 400.05], [179.95, 180, 180, 180.05]) = [179.95, 180, 180, 180.05] as given in Figure 3. This exhausts the fourth column by fuzzy zero, [-0.10, 0, 0, 0.10] and supply is reduced to ([399.95, 400, 400, 400.05] (-) [179.95, 180, 180, 180.05]) = [219.90, 220, 220, 220.10]. The fourth column is deleted from the Figure 3 such that we have the following shrunken matrix given in Figure 4. The values of penalties are also changed.

In Figure 4, the maximum value of Penalty is [7.9, 8, 8, 8.1] corresponding to $O_2$ row. In this allocation the cell having the fuzzy least cost is (2, 3). To this cell we allocate the minimum of Supply $A_2$ and Demand $B_3$ i.e., ([329.95, 330, 330, 330.05], [279.95, 280, 280, 280.05]) = [279.95, 280, 280, 280.05] as given in Figure 4. This exhausts the third column by fuzzy zero, [-0.10, 0, 0, 0.10] and supply is reduced to ([329.95, 330, 330, 330.05] (-) [279.95, 280, 280, 280.05]) = [49.90, 50, 50, 50.10]. The third column is deleted from the Figure 4 such that we have the following shrunken matrix given in Figure 5. As the third column is deleted the values of penalties are also changed.

|  |  | DESTINATIONS | | | | |
|---|---|---|---|---|---|---|
|  |  | $D_1$ | $D_3$ | $D_4$ | Supply ($A_i$) | Row Penalty |
| ORI | $O_1$ | [12.95, 13, 13, 13.05] | [15.95, 16, 16, 16.05] | [17.95, 18, 18, 18.05] | [29.90, 30, 30, 30.10] | [2.9, 3, 3, 3.1] |

| | | | | | |
|---|---|---|---|---|---|
| O₂ | [19.95, 20, 20, 20.05] | [10.95, 11, 11, 11.05] | [7.95, 8, 8, 8.05] | [329.95, 330, 330, 330.05] | [2.9, 3, 3, 3.1] |
| O₃ | [18.95, 19, 19, 19.05] | [16.95, 17, 17, 17.05] | **[10.95, 11, 11, 11.05]**<br>**([179.95, 180, 180, 180.05])** | [399.95, 400, 400, 400.05] | [5.9, 6, 6, 6.1] |
| Demand ($B_j$) | [299.95, 300, 300, 300.05] | [279.95, 280, 280, 280.05] | [179.95, 180, 180, 180.05]<br>**[-0.10, 0, 0, 0.10]** | | |
| Column Penalty | [5.9, 6, 6, 6.1] | [4.9, 5, 5, 5.1] | [2.9, 3, 3, 3.1] | | |

**Figure 3: 2nd Allocation to the Transportation Problem**

| | | DESTINATIONS | | | |
|---|---|---|---|---|---|
| | | $D_1$ | $D_3$ | Supply ($A_i$) | Row Penalty |
| **ORIGINS** | O₁ | [12.95, 13, 13, 13.05] | [15.95, 16, 16, 16.05] | [29.90, 30, 30, 30.10] | [2.9, 3, 3, 3.1] |
| | O₂ | [19.95, 20, 20, 20.05] | **[10.95, 11, 11, 11.05]**<br>**[279.95, 280, 280, 280.05]** | [329.95, 330, 330, 330.05] | [7.9, 8, 8, 8.1] |
| | O₃ | [18.95, 19, 19, 19.05] | [16.95, 17, 17, 17.05] | [219.90, 220, 220, 220.10] | [1.9, 2, 2, 2.1] |
| | Demand ($B_j$) | [299.95, 300, 300, 300.05] | [279.95, 280, 280, 280.05]<br>**[-0.10, 0, 0, 0.10]** | | |
| | Column Penalty | [5.9, 6, 6, 6.1] | [4.9, 5, 5, 5.1] | | |

**Figure 4: 3rd Allocation to the Transportation Problem**

In Figure 5, the maximum value of Penalty is [19.95, 20, 20, 20.05] corresponding to O₂ row. As there is only one element in this row, this cell i.e., (2, 1) is the fuzzy least cost cell and is considered for allocation. We allocate the minimum of Supply $A_2$ and Demand $B_1$ i.e., ([49.95, 50, 50, 50.05], [299.95, 300, 300, 300.05]) = [49.95, 50, 50, 50.05] as given in Figure 5. This exhausts the second row by fuzzy zero, [-0.10, 0, 0, 0.10] and demand is reduced to ([299.95, 300, 300, 300.05] (-) [49.95, 50, 50, 50.05]) = [249.90, 250, 250, 250.10]. The second row is deleted from the Figure 5 such that we have the following shrunken matrix given in Figure 6. As the second row is deleted the values of penalties are also changed.

| | | DESTINATIONS | | |
|---|---|---|---|---|
| | | $D_1$ | Supply ($A_i$) | Row Penalty |
| **ORIGINS** | O₁ | [12.95, 13, 13, 13.05] | [29.90, 30, 30, 30.10] | [12.95, 13, 13, 13.05] |
| | O₂ | **[19.95, 20, 20, 20.05]**<br>**[49.95, 50, 50, 50.05]** | [49.95, 50, 50, 50.05]<br>**[-0.10, 0, 0, 0.10]** | [19.95, 20, 20, 20.05] |
| | O₃ | [18.95, 19, 19, 19.05] | [219.90, 220, 220, 220.10] | [18.95, 19, 19, 19.05] |
| | Demand ($B_j$) | [299.95, 300, 300, 300.05] | | |
| | Column Penalty | [5.9, 6, 6, 6.1] | | |

**Figure 5: 4th Allocation to the Transportation Problem**

In Figure 6, the maximum value of Penalty is [18.95, 19, 19, 19.05] corresponding to O₃ row. As there is only one element in this row, this cell i.e., (3, 1) is the fuzzy least cost cell and is considered for allocation. We allocate the minimum of Supply $A_3$ and Demand $B_1$ i.e.,

([219.90, 220, 220, 220.10], [249.90, 250, 250, 250.10]) = [219.90, 220, 220, 220.10] as given in Figure 6. This exhausts the third row by fuzzy zero, [-0.10, 0, 0, 0.10] and demand is reduced to ([249.90, 250, 250, 250.10] (-) [219.90, 220, 220, 220.10]) = [29.85, 30, 30, 30.15]. The third row is deleted from the Figure 6 such that we have the following shrunken matrix given in Figure 7. As the third row is deleted the values of penalties are also changed.

|  |  | DESTINATIONS |  |  |
|---|---|---|---|---|
|  |  | $D_1$ | Supply ($A_i$) | Row Penalty |
| **ORIGINS** | $O_1$ | [12.95, 13, 13, 13.05] | [29.90, 30, 30, 30.10] | [12.95, 13, 13, 13.05] |
|  | $O_3$ | **[18.95, 19, 19, 19.05]** **[219.90, 220, 220, 220.10]** | [219.90, 220, 220, 220.10] **[-0.10, 0, 0, 0.10]** | [18.95, 19, 19, 19.05] |
|  | Demand ($B_j$) | [249.90, 250, 250, 250.05] |  |  |
|  | Column Penalty | [5.9, 6, 6, 6.1] |  |  |

**Figure 6: 5$^{th}$ Allocation to the Transportation Problem**

In Figure 7, only one cell value corresponding to $D_1$ column and $O_1$ row remains. There is only one Penalty value i.e., [12.95, 13, 13, 13.05]. In this allocation the cell having the fuzzy least cost is (1, 1). To this cell we allocate the minimum of Supply $A_1$ and Demand $B_1$ i.e., ([29.90, 30, 30, 30.10], [29.85, 30, 30, 30.15]) = [29.90, 30, 30, 30.10] as given in Figure 7. This exhausts the first row and first column by fuzzy zero, [-0.20, 0, 0, 0.20] and all the requirements are fulfilled in the 6$^{th}$ allocation. The demand and supply are differed by fuzzy zero ([-0.35, 0, 0, 0.35] (-) [-0.20, 0, 0, 0.20]) = [-0.55, 0, 0, 0.55] due to fuzziness. Finally, the initial basic feasible solution is shown in figure 8.

There are 6 positive independent allocations given by $m + n - 1 = 3 + 4 - 1$. This ensures that the solution is a fuzzy non – degenerate basic feasible solution. The total transportation cost = {$C_{11}$ (·) $X_{11}$} (+) {$C_{12}$ (·) $X_{12}$} (+) {$C_{21}$ (·) $X_{21}$} (+) {$C_{23}$ (·) $X_{23}$} (+){$C_{31}$ (·) $X_{31}$} (+) {$C_{34}$ (·) $X_{34}$} = {[12.95, 13, 13, 13.05] (·) [29.90, 30, 30, 30.10]} (+) {[14.95, 15, 15, 15.05] (·) [249.95, 250, 250, 250.05]} (+) {[19.95, 20, 20, 20.05] (·) [49.95, 50, 50, 50.05]} (+){[10.95, 11, 11, 11.05] (·) [279.95, 280, 280, 280.05]} (+) {[18.95, 19, 19, 19.05] (·) [219.90, 220, 220, 220.10]} (+) {[10.95, 11, 11, 11.05] (·) [179.95, 180, 180, 180.05]}
= [14323.47, 14380, 14380, 14436.57]                                            (9)

Here, $C_{ij}$ are cost coefficients and $X_{ij}$ are allocations (i = 1, 2, 3; j = 1, 2, 3, 4).

|  |  | DESTINATIONS |  |  |
|---|---|---|---|---|
|  |  | $D_1$ | Supply ($A_i$) | Row Penalty |
| **ORIGINS** | $O_1$ | **[12.95, 13, 13, 13.05]** **[29.90, 30, 30, 30.10]** | [29.90, 30, 30, 30.10] **[-0.60, 0, 0, 0.60]** | [12.95, 13, 13, 13.05] |
|  | Demand ($B_j$) | [29.85, 30, 30, 30.15] |  |  |
|  | Column Penalty | [12.95, 13, 13, 13.05] |  |  |

**Figure 7: 6$^{th}$ Allocation to the Transportation Problem**

|  |  | DESTINATIONS |  |  |  |
|---|---|---|---|---|---|
|  |  | $D_1$ | $D_2$ | $D_3$ | $D_4$ | Supply ($A_i$) |
| **ORIGI** | $O_1$ | **[12.95, 13, 13, 13.05]** ([29.90, 30, | **[14.95, 15, 15, 15.05]** ([249.95, 250, | [15.95, 16, 16, 16.05] | [17.95, 18, 18, 18.05] | [279.95, 280, 280, 280.05] |

| | | 30, 30.10]) | 250, 250.05]) | | | |
|---|---|---|---|---|---|---|
| | $O_2$ | [19.95, 20, 20, 20.05] ([49.95, 50, 50, 50.05]) | [21.95, 22, 22, 22.05] | **[10.95, 11, 11, 11.05]** ([279.95, 280, 280, 280.05]) | [7.95, 8, 8, 8.05] | [329.95, 330, 330, 330.05] |
| | $O_3$ | [18.95, 19, 19, 19.05] ([219.90, 220, 220, 220.10]) | [24.95, 25, 25, 25.05] | [16.95, 17, 17, 17.05] | **[10.95, 11, 11, 11.05]** ([179.95, 180, 180, 180.05]) | [399.95, 400, 400, 400.05] |
| **Demand ($B_j$)** | | [299.95, 300, 300, 300.05] | [249.95, 250, 250, 250.05] | [279.95, 280, 280, 280.05] | [179.95, 180, 180, 180.05] | |

**Figure 8: The final allocated matrix with the corresponding allocation values**

### 5.2 *Fuzzy Membership Functions of the costs and allocations*

Now, we present the fuzzy membership function of $C_{ij}$ and $X_{ij}$ and then the transportation costs.

$$\mu_{c_{11}}(x) = \begin{cases} (x-12.95)/0.05, 12.95 \le x \le 13 \\ 1, 13 \le x < 13 \\ (13.05-x)/0.05, 13 \le x \le 13.05 \\ 0, otherwise \end{cases}$$

Let $C_{11\alpha}$ be the interval of confidence for the level of presumption $\alpha$, $\alpha \in [0, 1]$.

Thus, $C_{11\alpha} = [c_1^{(\alpha)}, c_2^{(\alpha)}] = [0.05\alpha + 12.95, 13.05 - 0.05\alpha]$ \hfill (10)

$$\mu_{X_{11}}(x) = \begin{cases} (x-29.90)/0.10, 29.90 \le x \le 30 \\ 1, 30 \le x < 30 \\ (30.10-x)/0.10, 30 \le x \le 30.10 \\ 0, otherwise \end{cases}$$

Let $X_{11\alpha}$ be the interval of confidence for the level of presumption $\alpha$, $\alpha \in [0, 1]$.

Thus, $X_{11\alpha} = [x_1^{(\alpha)}, x_2^{(\alpha)}] = [0.10\alpha + 29.90, 30.10 - 0.10\alpha]$ \hfill (11)

$C_{11\alpha} (\cdot) X_{11\alpha} = [(0.05\alpha + 12.95)(0.10\alpha + 29.90), (13.05 - 0.05\alpha)(30.10 - 0.10\alpha)]$
$= [0.005\alpha^2 + 2.79\alpha + 387.205, 0.005\alpha^2 - 2.81\alpha + 392.805]$ \hfill (12)

Similarly, we can write the other fuzzy membership functions as follows:

$$\mu_{c_{12}}(x) = \begin{cases} (x-14.95)/0.05, 14.95 \le x \le 15 \\ 1, 15 \le x < 15 \\ (15.05-x)/0.05, 15 \le x \le 15.05 \\ 0, otherwise \end{cases}$$

$C_{12\alpha} = [0.05\alpha + 14.95, 15.05 - 0.05\alpha]$ \hfill (13)

$$\mu_{X_{12}}(x) = \begin{cases} (x - 249.95)/0.05, 249.95 \leq x \leq 250 \\ 1, 250 \leq x < 250 \\ (250.05 - x)/0.05, 250 \leq x \leq 250.05 \\ 0, otherwise \end{cases}$$

$$X_{12\alpha} = [0.05\alpha + 249.95, 250.05 - 0.05\alpha] \tag{14}$$

$$C_{12\alpha}(\cdot) X_{12\alpha} = [(0.05\alpha + 14.95)(0.05\alpha + 249.95), (15.05 - 0.05\alpha)(250.05 - 0.05\alpha)]$$
$$= [0.0025\alpha^2 + 13.245\alpha + 3736.7525, 0.0025\alpha^2 - 13.255\alpha + 3763.2525] \tag{15}$$

$$\mu_{c_{21}}(x) = \begin{cases} (x - 19.95)/0.05, 19.95 \leq x \leq 20 \\ 1, 20 \leq x < 20 \\ (20.05 - x)/0.05, 20 \leq x \leq 20.05 \\ 0, otherwise \end{cases}$$

$$C_{21\alpha} = [0.05\alpha + 19.95, 20.05 - 0.05\alpha] \tag{16}$$

$$\mu_{X_{21}}(x) = \begin{cases} (x - 49.95)/0.05, 49.95 \leq x \leq 50 \\ 1, 50 \leq x < 50 \\ (50.05 - x)/0.05, 50 \leq x \leq 50.05 \\ 0, otherwise \end{cases}$$

$$X_{21\alpha} = [0.05\alpha + 49.95, 50.05 - 0.05\alpha] \tag{17}$$

$$C_{21\alpha}(\cdot) X_{21\alpha} = [(0.05\alpha + 19.95)(0.05\alpha + 49.95), (20.05 - 0.05\alpha)(50.05 - 0.05\alpha)]$$
$$= [0.0025\alpha^2 + 3.495\alpha + 996.5025, 0.0025\alpha^2 - 3.505\alpha + 1003.5025] \tag{18}$$

$$\mu_{c_{23}}(x) = \begin{cases} (x - 10.95)/0.05, 10.95 \leq x \leq 11 \\ 1, 11 \leq x < 11 \\ (11.05 - x)/0.05, 11 \leq x \leq 11.05 \\ 0, otherwise \end{cases}$$

$$C_{23\alpha} = [0.05\alpha + 10.95, 11.05 - 0.05\alpha] \tag{19}$$

$$\mu_{X_{23}}(x) = \begin{cases} (x - 279.95)/0.05, 279.95 \leq x \leq 280 \\ 1, 280 \leq x < 280 \\ (280.05 - x)/0.05, 280 \leq x \leq 280.05 \\ 0, otherwise \end{cases}$$

$$X_{23\alpha} = [0.05\alpha + 279.95, 280.05 - 0.05\alpha] \tag{20}$$

$$C_{23\alpha}(\cdot) X_{23\alpha} = [(0.05\alpha + 10.95)(0.05\alpha + 279.95), (11.05 - 0.05\alpha)(280.05 - 0.05\alpha)]$$
$$= [0.0025\alpha^2 + 14.545\alpha + 3065.4525, 0.0025\alpha^2 - 14.555\alpha + 3094.5525] \tag{21}$$

$$\mu_{c_{31}}(x) = \begin{cases} (x-18.95)/0.05, 18.95 \le x \le 19 \\ 1, 19 \le x < 19 \\ (19.05-x)/0.05, 19 \le x \le 19.05 \\ 0, otherwise \end{cases}$$

$$C_{31\alpha} = [0.05\alpha + 18.95, 19.05 - 0.05\alpha] \tag{22}$$

$$\mu_{X_{31}}(x) = \begin{cases} (x-219.90)/0.10, 219.90 \le x \le 220 \\ 1, 220 \le x < 220 \\ (220.10-x)/0.10, 220 \le x \le 220.10 \\ 0, otherwise \end{cases}$$

$$X_{31\alpha} = [0.10\alpha + 219.90, 220.10 - 0.10\alpha] \tag{23}$$

$$C_{31\alpha}(\cdot) X_{31\alpha} = [(0.05\alpha + 18.95)(0.10\alpha + 219.90), (19.05 - 0.05\alpha)(220.10 - 0.10\alpha)]$$
$$= [0.005\alpha^2 + 12.89\alpha + 4167.105, 0.005\alpha^2 - 12.91\alpha + 4192.905] \tag{24}$$

$$\mu_{c_{34}}(x) = \begin{cases} (x-10.95)/0.05, 10.95 \le x \le 11 \\ 1, 11 \le x < 11 \\ (11.05-x)/0.05, 11 \le x \le 11.05 \\ 0, otherwise \end{cases}$$

$$C_{34\alpha} = [0.05\alpha + 10.95, 11.05 - 0.05\alpha] \tag{25}$$

$$\mu_{X_{34}}(x) = \begin{cases} (x-179.95)/0.05, 179.95 \le x \le 180 \\ 1, 180 \le x < 180 \\ (180.05-x)/0.05, 180 \le x \le 180.05 \\ 0, otherwise \end{cases}$$

$$X_{34\alpha} = [0.05\alpha + 179.95, 180.05 - 0.05\alpha] \tag{26}$$

$$C_{34\alpha}(\cdot) X_{34\alpha} = [(0.05\alpha + 10.95)(0.05\alpha + 179.95), (11.05 - 0.05\alpha)(180.05 - 0.05\alpha)]$$
$$= [0.0025\alpha^2 + 9.545\alpha + 1970.4525, 0.0025\alpha^2 - 9.555\alpha + 1989.5525] \tag{27}$$

Now, the total transportation cost is given by: $\text{Cost}_\alpha = \{C_{11\alpha}(\cdot) X_{11\alpha}\} (+) \{C_{12\alpha}(\cdot) X_{12\alpha}\} (+) \{C_{21\alpha}(\cdot) X_{21\alpha}\} (+) \{C_{23\alpha}(\cdot) X_{23\alpha}\} (+) \{C_{31\alpha}(\cdot) X_{31\alpha}\} (+) \{C_{34\alpha}(\cdot) X_{34\alpha}\}$
$$= [0.02\alpha^2 + 56.51\alpha + 14323.47, 0.02\alpha^2 - 53.78\alpha + 14436.57] \tag{28}$$

The expression given in equation (19) is obtained using the equations (12), (15), (18), (21), (24) and (27). From equation (28) we solve the following equations whose roots $\in [0, 1]$:

$$0.02\alpha^2 + 56.51\alpha + 14323.47 - x_1 = 0 \tag{29}$$

$$0.02\alpha^2 - 53.78\alpha + 14436.57 - x_2 = 0 \tag{30}$$

From equation (29) we have,

$$\alpha = \{-56.51 + \sqrt{((56.51)^2 - 4 \times 0.02 \times (14323.47 - x_1))}\}/(2 \times 0.02)$$

From equation (30) we have,

$$\alpha = \{53.78 - \sqrt{((53.78)^2 - 4 \times 0.02 \times (14436.57 - x_2))}\}/(2 \times 0.02)$$

$$\mu_{\cos t}(x) = \begin{cases} \{-56.51 + \sqrt{((56.51)^2 - 4 \times 0.02 \times (14323.47 - x_1))}\}/(2 \times 0.02), 14323.47 \leq x \leq 14380 \\ 1, 14380 \leq x < 14380 \\ \{53.78 - \sqrt{((53.78)^2 - 4 \times 0.02 \times (14436.57 - x_2))}\}/(2 \times 0.02), 14380 \leq x \leq 14436.57 \\ 0, otherwise \end{cases}$$

which is the required fuzzy membership function of the transportation cost (using equation (9)).

### 5.3 *Optimality Test by FMODIM*

To determine the fuzzy optimal solution for the above problem, we make use of the FMODIM. We determine the set of numbers $U_i$ and $V_j$ for each row and column of the matrix given in figure 8 with $U_i$ (+) $V_j = C_{ij}$ for each occupied cell. We assign the value of *fuzzy zero* to $U_1 = [-0.05, 0, 0, 0.05]$ arbitrarily, as all the rows have the identical number of allocations. From the occupied cells we have,

$V_1 = C_{11}$ (-) $U_1 = [12.95, 13, 13, 13.05]$ (-) $[-0.05, 0, 0, 0.05] = [12.90, 13, 13, 13.10]$
$V_2 = C_{12}$ (-) $U_1 = [14.95, 15, 15, 15.05]$ (-) $[-0.05, 0, 0, 0.05] = [14.90, 15, 15, 15.10]$
$U_2 = C_{21}$ (-) $V_1 = [19.95, 20, 20, 20.05]$ (-) $[12.90, 13, 13, 13.10] = [6.85, 7, 7, 7.15]$
$V_3 = C_{23}$ (-) $U_2 = [10.95, 11, 11, 11.05]$ (-) $[6.85, 7, 7, 7.15] = [3.80, 4, 4, 4.20]$
$U_3 = C_{31}$ (-) $V_1 = [18.95, 19, 19, 19.05]$ (-) $[12.90, 13, 13, 13.10] = [5.85, 6, 6, 6.15]$
$V_4 = C_{34}$ (-) $U_3 = [10.95, 11, 11, 11.05]$ (-) $[5.85, 6, 6, 6.15] = [4.80, 5, 5, 5.20]$

Now we calculate the sum of $U_i$ and $V_j$ for each unoccupied cell. The values of $U_i$ (+) $V_j$ are given below the $C_{ij}$ value of the cells which are as follows:

$U_1$ (+) $V_3 = [-0.05, 0, 0, 0.05]$ (+) $[3.80, 4, 4, 4.20] = [3.75, 4, 4, 4.25]$
$U_1$ (+) $V_4 = [-0.05, 0, 0, 0.05]$ (+) $[4.80, 5, 5, 5.20] = [4.75, 5, 5, 5.25]$
$U_2$ (+) $V_2 = [6.85, 7, 7, 7.15]$ (+) $[14.90, 15, 15, 15.10] = [21.75, 22, 22, 22.25]$
$U_2$ (+) $V_4 = [6.85, 7, 7, 7.15]$ (+) $[4.80, 5, 5, 5.20] = [11.65, 12, 12, 12.35]$
$U_3$ (+) $V_2 = [5.85, 6, 6, 6.15]$ (+) $[14.90, 15, 15, 15.10] = [20.75, 21, 21, 21.25]$
$U_3$ (+) $V_3 = [5.85, 6, 6, 6.15]$ (+) $[3.80, 4, 4, 4.20] = [9.65, 10, 10, 10.35]$

Next we find the net evaluations $\Delta_{ij} = C_{ij}$ (-) ($U_i$ (+) $V_j$) for each unoccupied cell. The values of $\Delta_{ij}$ are given below the values of $U_i$ (+) $V_j$ which are as follows:

$\Delta_{13} = C_{13}$ (-) ($U_1$ (+) $V_3$) = [15.95, 16, 16, 16.05] (-)
 ([-0.05, 0, 0, 0.05] (+) [3.80, 4, 4, 4.20]) = [11.80, 12, 12, 12.20]
$\Delta_{14} = C_{14}$ (-) ($U_1$ (+) $V_4$) = [17.95, 18, 18, 18.05] (-)
 ([-0.05, 0, 0, 0.05] (+) [4.75, 5, 5, 5.25]) = [12.75, 13, 13, 13.25]
$\Delta_{22} = C_{22}$ (-) ($U_2$ (+) $V_2$) = [21.95, 22, 22, 22.05] (-)
 ([6.85, 7, 7, 7.15] (+) [14.90, 15, 15, 15.10]) = [-0.20, 0, 0, 0.20]
$\Delta_{24} = C_{24}$ (-) ($U_2$ (+) $V_4$) = [7.95, 8, 8, 8.05] (-)
 ([6.85, 7, 7, 7.15] (+) [4.80, 5, 5, 5.20]) = [3.70, 4, 4, 4.30]
$\Delta_{32} = C_{32}$ (-) ($U_3$ (+) $V_2$) = [24.95, 25, 25, 25.05] (-)
 ([9.65, 10, 10, 10.35] (+) [14.90, 15, 15, 15.10]) = [-0.40, 0, 0, 0.40]
$\Delta_{33} = C_{33}$ (-) ($U_3$ (+) $V_3$) = [16.95, 17, 17, 17.05] (-)
 ([9.65, 10, 10, 10.35] (+) [3.80, 4, 4, 4.20]) = [2.50, 3, 3, 3.50]

The above calculated values are given in figure 9. Here, as all $\Delta_{ij} \geq [-0.40, 0, 0, 0.40]$ the solution obtained is optimal in nature and an alternative solution exists given by $\Delta_{32} = [-0.40, 0, 0, 0.40]$. Therefore, the optimal allocations are given by the following:

$$X_{11} = [29.90, 30, 30, 30.10]; X_{12} = [249.95, 250, 250, 250.05]$$
$$X_{21} = [49.95, 50, 50, 50.05]; X_{23} = [279.95, 280, 280, 280.05]$$
$$X_{31} = [219.90, 220, 220, 220.10]; X_{34} = [179.95, 180, 180, 180.05]$$

The total optimum transportation cost = [14323.47, 14380, 14380, 14436.57]

The fuzzy membership functions of the optimality test can be obtained using similar technique as in the case for the fuzzy initial basic feasible solution.

## 6. Computational Complexity

Here, we investigate the computational complexity of the transportation problem using Vogel's Approximation Method with $m$ origins and $n$ destinations. Let the total computational time be given by T $(m, n)$. The time to calculate the penalty values for $m$ rows $= (n + 1) m$. The time to calculate the penalty values for $n$ columns $= (m + 1) n$. The time to search for the maximum value of the penalty for the corresponding least value of the cost $= \{m / (m + n)\} n$, if maximum penalty is found in a row and the least value of the cost $= \{n / (m + n)\} m$, if maximum penalty is found in a column.

Now, the time required to obtain the feasible solution corresponding to $(m + n - 1)$ cell allocations

$$= \begin{cases} \dfrac{m}{(m+n)} n(m+n-1) \\ \dfrac{n}{(m+n)} m(m+n-1) \end{cases}.$$

Hence, the total computational time required is given by,

$$\text{T}(m, n) = \begin{cases} (n+1)m + \dfrac{m}{(m+n)} n(m+n-1) \\ (m+1)n + \dfrac{n}{(m+n)} m(m+n-1) \end{cases}$$

Thus, the total time complexity is O $(mn)$. The computational time grows as the values of $m$ and $n$ increases, as a result of which the problem becomes intractable in nature.

| | | DESTINATIONS | | | | |
|---|---|---|---|---|---|---|
| | | $D_1$ | $D_2$ | $D_3$ | $D_4$ | $U_i$ |
| ORIGINS | $O_1$ | **[12.95, 13, 13, 13.05]** ([29.90, 30, 30, 30.10]) | **[14.95, 15, 15, 15.05]** ([249.95, 250, 250, 250.05]) | [15.95, 16, 16, 16.05] [3.75, 4, 4, 4.25] [11.80, 12, 12, 12.20] | [17.95, 18, 18, 18.05] [4.75, 5, 5, 5.25] [12.75, 13, 13, 13.25] | [-0.05, 0, 0, 0.05] |
| | $O_2$ | **[19.95, 20, 20, 20.05]** ([49.95, 50, 50, 50.05]) | [21.95, 22, 22, 22.05] [21.75, 22, 22, 22.25] | **[10.95, 11, 11, 11.05]** ([279.95, 280, 280, 280.05]) | [7.95, 8, 8, 8.05] [11.65, 12, 12, 12.35] | [6.85, 7, 7, 7.15] |

|   |   |   | [-0.20, 0, 0, 0.20] |   | [3.70, 4, 4, 4.30] |   |
|---|---|---|---|---|---|---|
|   | $O_3$ | [18.95, 19, 19, 19.05] ([219.90, 220, 220, 220.10]) | [24.95, 25, 25, 25.05] [20.75, 21, 21, 21.25] [-0.40, 0, 0, 0.40] | [16.95, 17, 17, 17.05] [9.65, 10, 10, 10.35] [2.50, 3, 3, 3.50] | **[10.95, 11, 11, 11.05]** ([179.95, 180, 180, 180.05]) | [5.85, 6, 6, 6.15] |
|   | $V_j$ | [12.90, 13, 13, 13.10] | [14.90, 15, 15, 15.10] | [3.80, 4, 4, 4.20] | [4.80, 5, 5, 5.20] |   |

**Figure 9:** The final allocated matrix with the sum of the $U_i$ and $V_j$ values and the cell evaluations for the unoccupied cells

## 7. Discussions

The degeneracy in transportation problem and unbalanced transportation problem can be similarly represented using fuzzy trapezoidal numbers. This model of fuzzy trapezoidal numbers for the transportation problem can be easily extended to the assignment problem which is a special class of the transportation problem where the number of origins is equal to the number of destinations. However, if the FVAM algorithm for obtaining the solution of transportation problem is applied to solve the assignment problem a large number of iterations have to be performed for the resolution of degeneracy till the optimal solution is obtained. In assignment problem it is observed that a basic feasible solution for the constraint equations will consist of ($2m$ - 1) variables. But it is observed that every basic solution will consist of $m$ basic variables equal to 1 and ($m - 1$) basic variables equal to 0 and as such the basic feasible solution will have a high level of degeneracy [5], [12]. This will increase the overall computational complexity of the solution. The effectiveness of the solutions obtained for the transportation problem can greatly be enhanced by incorporating the genetic algorithms alongwith the fuzzy trapezoidal numbers such that the computational complexity is greatly reduced.

## 8. Conclusions

This Paper presents the closed, bounded and non–empty feasible region of the transportation problem using fuzzy trapezoidal numbers which ensures the existence of an optimal solution to the balanced transportation problem. The multi-valued nature of Fuzzy Sets allows handing of uncertainty and vagueness involved in the cost values of each cells in the transportation table. For finding the initial solution of the transportation problem we use the FVAM and for determining the optimality of the obtained solution FMODIM is used. The fuzzification of the cost of the transportation problem is discussed with the help of a numerical example. We also discuss the computational complexity involved in the problem. The effectiveness of the solutions obtained for the problem can greatly be enhanced by incorporating the genetic algorithms alongwith the fuzzy trapezoidal numbers such that the computational complexity is greatly reduced.